\documentclass{article}
\usepackage{spconf,amsmath,graphicx}

\usepackage{booktabs}
\usepackage{cleveref}
\usepackage{xcolor,colortbl}
\usepackage{stfloats}
\usepackage{multirow}
\usepackage[hyphens]{url}
\definecolor{Gray}{gray}{0.95}

\usepackage[utf8]{inputenc}
\usepackage[english]{babel}
\usepackage{fancyhdr}

\title{Hint-enhanced In-Context Learning wakes Large Language Models up for knowledge-intensive tasks}
\name{Yifan Wang$^{1}$, Qingyan Guo$^{1}$, Xinzhe Ni$^{1}$, Chufan Shi$^{1}$, Lemao Liu$^{2}$, Haiyun Jiang$^{2\dagger}$, Yujiu Yang$^{1\dagger}\thanks{$^\dagger$Corresponding author.}$}
\address{$^1$ Tsinghua Shenzhen International Graduate School, Tsinghua University \quad $^2$ Tencent AI Lab\\
\{wangyifa22, gqy22, nxz22, scf22\}@mails.tsinghua.edu.cn,\\
\{redmondliu, haiyunjiang\}@tencent.com,\quad yang.yujiu@sz.tsinghua.edu.cn
}

\begin{document}
%

\maketitle

\begin{abstract}

In-context learning (ICL) ability has emerged with the increasing scale of large language models (LLMs), enabling them to learn input-label mappings from demonstrations and perform well on downstream tasks. 
However, under the standard ICL setting, LLMs may sometimes neglect query-related information in demonstrations, leading to incorrect predictions. 
To address this limitation, we propose a new paradigm called Hint-enhanced In-Context Learning (HICL) to explore the power of ICL in open-domain question answering, an important form in knowledge-intensive tasks.
HICL leverages LLMs' reasoning ability to extract query-related knowledge from demonstrations, then concatenates the knowledge to prompt LLMs in a more explicit way. 
Furthermore, we track the source of this knowledge to identify specific examples, and introduce a Hint-related Example Retriever (HER) to select informative examples for enhanced demonstrations. 
We evaluate HICL with HER on 3 open-domain QA benchmarks, and observe average performance gains of 2.89 EM score and 2.52 F1 score on gpt-3.5-turbo, 7.62 EM score and 7.27 F1 score on LLaMA-2-Chat-7B compared with standard setting.


\end{abstract}

\section{Introduction}
\label{sec:intro}

Large language models (LLMs) with in-context learning (ICL) ability have attracted wide attention. Prompted by demonstrations consisting of a few input-label pairs, LLMs perform well even on unseen tasks \cite{brown2020language, 2023arXivSurvey}.
ICL ability strongly depends on selected training examples \cite{liu2021makes,2021arXivCalibrate}, and some methods \cite{gonen2022demystifying, guo2023connecting} are designed to form a high-quality demonstration.
Recent works have explored what makes ICL work in LLMs.
Following the format of examples in demonstrations, language models can predict the right labels by utilizing prior knowledge acquired from pretraining~\cite{min2022rethinking, Shisg}.
As the model scale further increases, LLMs can acquire knowledge directly from input-label mappings in demonstrations \cite{wei2023larger}. 

\begin{figure*}[t]
   \centering
   \includegraphics[width=0.85\linewidth]{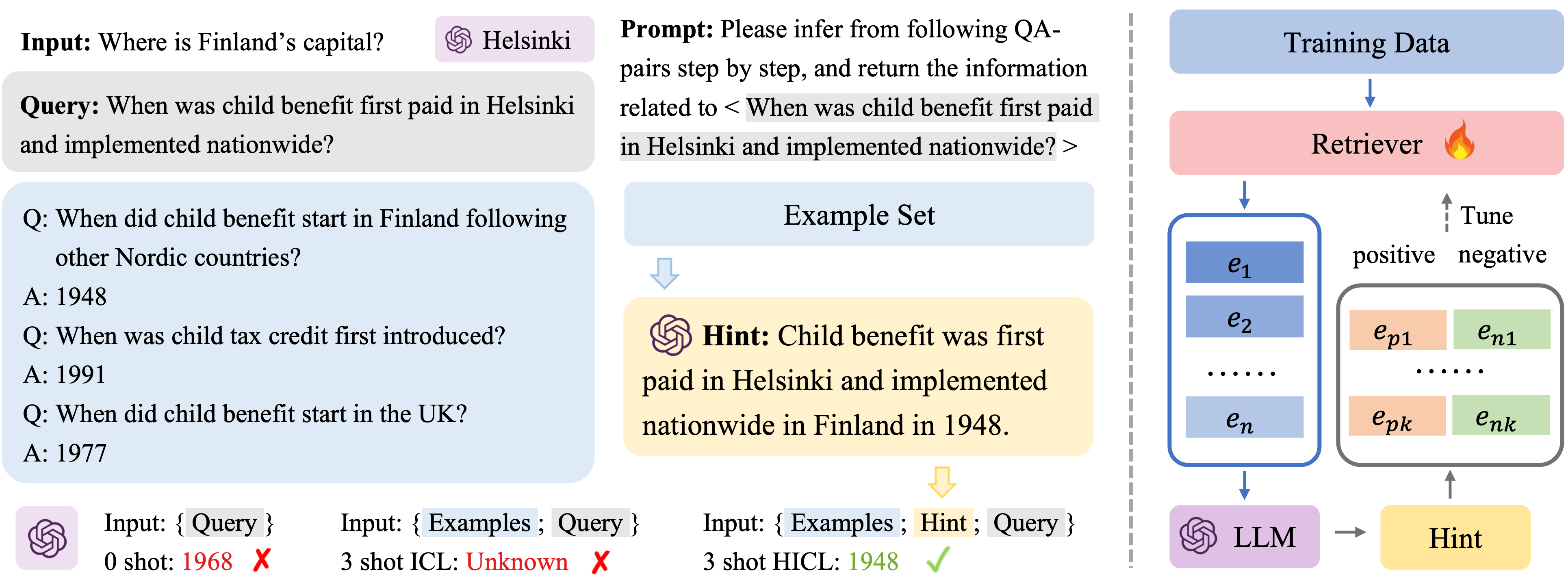}
   \caption{Left shows frames of standard ICL and HICL. We conduct experiments on gpt-3.5-turbo. While LLM neglects relevant knowledge under standard setting, it can predict correctly after concatenating the hint. Right is an overview of training HER. We use a retriever to obtain an example set $E$ for each query. Then LLM will follow the prompt and summarize query-related hints from $E$. Based on the overlap with hints, we can label examples with positive or negative for tuning.}
   \label{fig:HICL}
\end{figure*}

However, we found that the ICL ability of LLMs is limited in knowledge-intensive tasks, using the input form $\{x_1, y_1, \ldots, x_k, y_k; query\}$ as standard ICL setting. 
As shown in Figure~\ref{fig:HICL}, firstly, we verified that LLM knows \emph{[\underline{Helsinki} is \underline{Finland}'s capital]} indeed. Nevertheless, when coming to the question \emph{[When was child benefit first paid in \underline{Helsinki} and implemented nationwide?]}, LLM still has a certain probability of neglecting the knowledge in \emph{[Q: When did child benefit start in \underline{Finland} following other Nordic countries? A: 1948]}.
To obtain a universal conclusion, we conduct a \textbf{pilot experiment}\footnote{We sample 600 queries from the test set of Natural Questions \cite{kwiatkowski2019natural} and experimented with Dense Passage Retrieval \cite{karpukhin2020dense} to select examples.\label{pilot_exp}}. Under 5-shot setting, we found that 42.2\% of test queries contain corresponding knowledge in demonstrations, but LLM with standard ICL only predicts correctly for 69.9\% in this subset.
To better exploit the information contained in demonstrations, we propose a new paradigm called \textbf{H}int enhanced \textbf{I}n-\textbf{C}ontext \textbf{L}earning (HICL).

Firstly, we generate hints to present query-related information in a more \emph{explicit} way. 
Prior works \cite{Self-Ask, DSP} investigate the ability of LLMs to perform compositional reasoning tasks in multi-hop question answering tasks. With structured prompts, we can further explore the potential of frozen LLMs.
Hence, we propose a prompt-based knowledge-extracting scheme to extract valuable information from selected examples, leveraging LLMs' reasoning and summarizing ability \cite{2022arXivCoT,2023arXivSummarization}. 
As shown in Figure~\ref{fig:HICL}, LLM generates the hint \emph{[Child benefit was first paid in Helsinki and implemented nationwide in Finland in 1948.]} based on related examples.
In this way, the informative knowledge contained in demonstrations will be explored fully by the LLM, which helps LLM to answer the question consequently. 

With hints generated by the LLM, we can locate specific examples that provide knowledge. 
But in some cases, hint-related examples are not the closest to the test queries. 
We further analyze the \textbf{pilot experiment}\footref{pilot_exp} and locate specific examples that provide knowledge. Nearly 21\% of hint-related examples are not ranked top-1 when we can extract hints under 5-shot setting.
It turns out that the example selection method still has room for improvement, and we are inspired to propose a Hint-related Example Retriever (HER). 
We use hints as supervisory information to label examples as positive or negative by calculating the similarity score, then train a retriever by contrastive learning from the data. This paradigm can also be applied to Black-Box LLMs, where the output probability distribution is not accessible.

To conclude, the main contributions of our work are as follows: 
(1) HICL effectively enhances ICL ability in knowledge-intensive tasks with hints generated by LLMs. Experiments show that our proposed paradigm improves the performance on 3 open-domain QA benchmarks, where LLMs may fail to leverage relevant knowledge in demonstrations.
(2) To retrieve hint-related examples better, we propose a new example selection model called Hint-related Example Retriever (HER). Through HER, we can retrieve examples with query-related knowledge more effectively. 

\section{Method}
In this section, we will present the definition of HICL into two parts: (1) Hint Extraction Module for getting query-related knowledge in Section~\ref{section:HEM}; (2) Hint-related Example Retriever for selecting optimal examples in Section~\ref{section:HER}.

\subsection{Hint Extraction Module}
\label{section:HEM}
To prompt LLMs in a more explicit way, we utilize LLMs' reasoning ability to generate hints containing query-related knowledge. 
Firstly, we take advantage of a retriever to obtain an example set $E = \{ e_{1}, \ldots, e_{k}\}$, where $e_{n}=\{x_{n}, y_{n}\}, n\in [1,k]$ represents the $n$-th similar example to the corresponding query.
We prompt LLMs to extract hints via the following instruction: \emph{Please infer from the following QA-pairs step by step, and return the information related to \texttt{[query]}. If there is no information, please return ''None''}. 
Then LLMs return query-related information by extracting from the examples retrieved. To avoid noises, we filtered out responses without knowledge, namely, the cases where LLM returns ''None''. 
We append the hint before the test query, prompting LLMs to generate the final answer under the few-shot setting.
In this way, LLMs can further explore the latent knowledge in $E$ and achieve better performance.

\subsection{Hint-related Example Retriever}
\label{section:HER}

After obtaining hints from $E$, we can locate the specific examples that provide relevant knowledge.
However, in some cases, hint-related examples are not the closest to test queries.
This inspired us to propose a Hint-related Example Retriever for selecting examples with relevant knowledge more effectively, by further exploiting hints with contrastive learning, as illustrated in Figure~\ref{fig:HICL}. 

\textbf{Training Data Construction} \quad 
To build training data based on the hints, we sample queries from the Natural Questions \cite{kwiatkowski2019natural} training set and retrieve top-k similar examples to extract hints. Hence, we can obtain an example set $E_i$ and hint $h_i$ for query $x_i$.
According to the overlap with $h_i$, we consider the example with a high F1 score as the hint-related example, namely, the positive one. And we sample the negative example $e_{i}^{-}$ with a low F1 score from the rest set of $E_i$. Then we obtain triplets as the form $\{x_{i}; e_{i}^{+}; e_{i}^{-}\}$.
We retrieved top-10 similar examples based on the RoBERTa*, an unsupervised model fine-tuned on the RoBERTa-large model with SNLI, MultiNLI, and STS-B datasets\cite{bowman2015large,williams2017broad,cer2017semeval}. We extracted hints by gpt-3.5-turbo\footnote{https://chat.openai.com\label{chatgpt}} from demonstrations and labeled positive and negative examples for each query. Finally, we got 3429 triples as the training data.

\begin{table*}
\setlength{\tabcolsep}{10pt}
\centering
\resizebox{1.87\columnwidth}{!}
{
\begin{tabular}{llllllll|ll}
\toprule
& &\multicolumn{2}{c}{\textbf{NQ}}& \multicolumn{2}{c}{\textbf{WebQ}} & \multicolumn{2}{c}{\textbf{TriviaQA}}& \multicolumn{2}{c}{\textbf{AVG}}\\
\hline
\textbf{Model} &\textbf{Method}& \multicolumn{1}{c}{\textbf{EM}} & \multicolumn{1}{c}{\textbf{F1}} & \multicolumn{1}{c}{\textbf{EM}} & \multicolumn{1}{c}{\textbf{F1}} & \multicolumn{1}{c}{\textbf{EM}} & \multicolumn{1}{c}{\textbf{F1}}& \multicolumn{1}{c}{\textbf{EM}} & \multicolumn{1}{c}{\textbf{F1}}\\
\hline 
\multirow{5}*{gpt-3.5-turbo}
& Zero-shot & 27.80 & 38.85 & 23.53 & 36.50 & 65.54 & 71.81 & 38.96 & 49.05 \\
& Standard ICL $_\mathrm{DPR}$ & 38.66 & 50.38 & 39.80 & 46.40 & 69.20 & 75.14 & 49.22 & 57.30 \\
& RECITE $_\mathrm{DPR}$ & 36.13 & 50.82 & 32.60 & 44.60 & 67.47 & 75.43 & 45.40 & 56.95 \\
& HICL $_\mathrm{DPR}$ & \underline{40.66} & \underline{51.41} & \underline{42.73} & \underline{48.80} & \textbf{70.93} & \textbf{76.64} & \underline{51.44} & \underline{58.95} \\
& HICL $_\mathrm{HER_{DPR}}$ & \textbf{41.53} & \textbf{52.75} & \textbf{43.94} & \textbf{50.12} & \underline{70.87} & \underline{76.58} & \textbf{52.11} & \textbf{59.82} \\
\hline 
\multirow{5}*{LLaMA-2-Chat-7B}
& Zero-shot & 4.33 & 14.06 & 9.27 & 23.00 & 24.00 & 33.41 & 12.53 & 23.49 \\
& Standard ICL $_\mathrm{DPR}$ & 18.53 & 29.23 & 22.40 & 34.18 & 38.80 & 45.33 & 26.58 & 36.25 \\
& RECITE $_\mathrm{DPR}$ & 16.80 & 28.46 & 24.80 & 37.85 & 43.87 & 51.77 & 28.49 & 39.36 \\
& HICL $_\mathrm{DPR}$ & \underline{25.07} & \underline{35.46} & \underline{28.07} & \textbf{39.84} & \underline{47.47} & \underline{53.40} & \underline{33.53} & \underline{42.90} \\
& HICL $_\mathrm{HER_{DPR}}$ & \textbf{25.60} & \textbf{36.38} & \textbf{28.33} & \underline{39.64} & \textbf{48.66} & \textbf{54.53} & \textbf{34.20} & \textbf{43.52} \\

\bottomrule
\end{tabular}
}
\caption{\label{tab:main-result}
Main results on 3 open-domain QA benchmarks. The indexes indicate retrieval models for selecting examples. We use 5 examples for the ICL demonstration setting. The best results are in bold, while the second best are underlined.
}
\end{table*}

\textbf{Training} \quad We train HER from labeled data by contrastive learning. In order to maximize the similarity score for the hint-related example with the query while minimizing the score for the negative example with the query, we use InfoNCE loss \cite{oord2018representation} to train the retriever. 
To compute loss, we structure the training data as $\{x_{i}; e_{i}^{+}; e_{i, 1}^{-}, \ldots e_{i, B}^{-}\}$, where $B$ denotes the training batch size.
In detail, negative examples contain (1) $1$ negative example from the same triplet, paired with the current query;~(2) $B-1$ negative examples paired with other queries. The loss function is defined as:

\begin{equation}
\begin{aligned}
& L\left(x_{i}, e_{i}^{+}, e_{i, 1}^{-}, \ldots e_{i, B}^{-}\right) \\
= & -\log \frac{e^{\operatorname{sim}\left(x_{i}, e_{i}^{+}\right)}}{e^{\operatorname{sim}\left(x_{i}, e_{i}^{+}\right)}+\sum_{j=1}^{B} e^{\operatorname{sim}\left(x_{i}, e_{i, j}^{-}\right)}}
\end{aligned}
\end{equation}

\section{Experiments and Results}

We apply our proposed method on three QA benchmarks: Natural Questions (NQ), Web Questions (WebQ), and TriviaQA \cite{kwiatkowski2019natural,berant2013semantic,joshi2017triviaqa}. 
We sample 300 queries each time as the test set and use Exact Match (EM) and F1 score to evaluate the overlap between the predicted label and ground truth. The mean scores over 5 random seeds are reported. 
We compare our HICL with standard ICL and few-shot RECITE \cite{recite}, which generate recite passages as hints from LLMs’ own memory before producing final answers. 

To further analyze the gains brought by HER, our work adopts RoBERTa* and DPR \cite{karpukhin2020dense} as the base semantic similarity models, and tunes them with hint-selected examples. DPR is a set of models for state-of-the-art open-domain QA research, which is widely used in retrieval-augmented generation (RAG) \cite{RAG} in knowledge-intensive tasks. We use the question encoder of DPR which is trained on NQ dataset.
To train HERs, We use the Adam optimizer \cite{2014Adam} with batch size 32 and learning rate 1e-5, based on RoBERTa* and DPR. 
We train for 5 epochs and save the checkpoint with the best retrieval performance. The cosine distance is employed to get the similarity score when we select ICL examples.

\subsection{Main results}
We report the results for gpt-3.5-turbo\footref{chatgpt} and LLaMA-2-Chat-7B\cite{llama2} in Table~\ref{tab:main-result}. The indexes indicate the retrieval model for select examples, and
HER$_\mathrm{DPR}$ refers to the model after being tuned by the hint-selected data as described in Section~\ref{section:HER}. 
It turns out that HICL outperforms other ICL settings on all the benchmarks with the same examples in few-shot setting. Moreover, after tuning the retriever with hint-selected examples from the training set, HICL$_\mathrm{HER{_{DPR}}}$ can achieve average performance gains of 2.89 EM score and 2.52 F1 score on gpt-3.5-turbo, 7.62 EM score and 7.27 F1 score on LLaMA-2-Chat-7B compared with standard setting.


\begin{table}
\setlength{\tabcolsep}{9pt}
\centering
\resizebox{0.85\columnwidth}{!}
{
\begin{tabular}{lllll}
\toprule
&\multicolumn{2}{c}{\textbf{Standard}}& \multicolumn{2}{c}{\textbf{HICL}}\\
\hline
\textbf{Model}& \multicolumn{1}{c}{\textbf{EM}} & \multicolumn{1}{c}{\textbf{F1}} & \multicolumn{1}{c}{\textbf{EM}} & \multicolumn{1}{c}{\textbf{F1}}\\
\hline 
RoBERTa* & 48.69 & 56.89 & 49.51 & 57.96 \\
 \hspace{4mm}+ HER & \textbf{49.89} & \textbf{57.88} & \textbf{51.98} & \textbf{59.43} \\
\hline
DPR & 49.22 & 57.30 & 51.44 & 58.95 \\
 \hspace{4mm}+ HER & \textbf{49.96} & \textbf{58.26} & \textbf{52.11} & \textbf{59.82} \\
\bottomrule
\end{tabular}
}
\caption{\label{tab:HER_type}
Analysis of HERs. "+HER" indicates tuning with hint-selected data on the base model. We report the average scores of gpt-3.5-turbo on 3 open-domain QA benchmarks.
}
\end{table}

\subsection{Effect of HER}
To explore whether selected examples are beneficial for each model, we conduct HER based on RoBERTa* and DPR. Table~\ref{tab:HER_type} reports the results. Under HICL setting, HERs achieve performance gains of 2.47 EM score and 1.47 F1 score on RoBERTa*, 0.67 EM score and 0.87 F1 score on DPR. 
Note that DPR question encoder has already been finetuned on NQ dataset before training, which may be the reason for HER bringing more performance gains on RoBERTa* than DPR.

\begin{figure}[t]
\setlength{\tabcolsep}{10pt}
   \centering
   \includegraphics[width=0.85\linewidth]{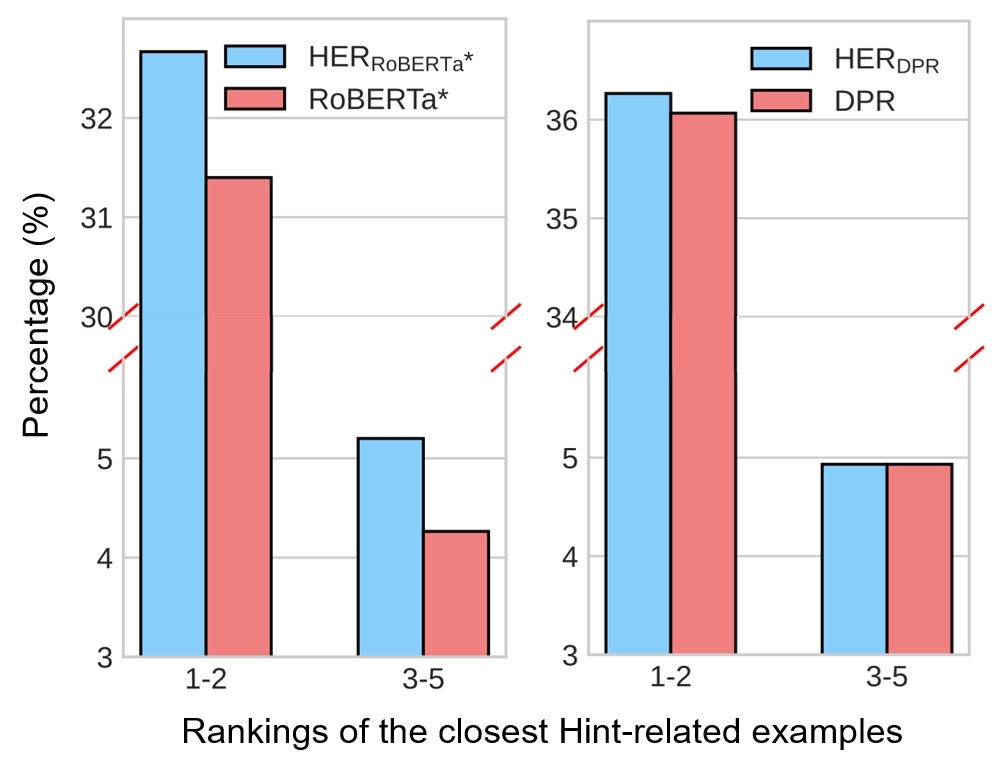}
   \caption{A statistics of rankings of the closest hint-related examples, e.g. for a certain query, a second position of the closest hint-related example will be counted into x-axis 1-2.}
   \label{fig:position}
\end{figure}

Moreover, we conducted an experiment on rankings of the closest hint-related examples.
Firstly, we set 5 random seeds and sampled 300 queries each time on NQ dataset. 
For each query, we select and order examples by similarity computation with different retrieval models. 
Then, we statistics positions of the closest hint-related examples.
As shown in Figure~\ref{fig:position}, more hint-related examples can be ranked higher in similarity with our HERs, tuned by the data we build in Section~\ref{section:HER}.
When retrieving the most similar $k$ examples to form a demonstration, LLM can generate more hints for the same few-shot setting. Our results demonstrate the similarity calculated by HERs can reflect knowledge correlation better.

\subsection{Number of Examples}
\begin{figure}[t]
   \centering
   \includegraphics[width=0.88\linewidth]{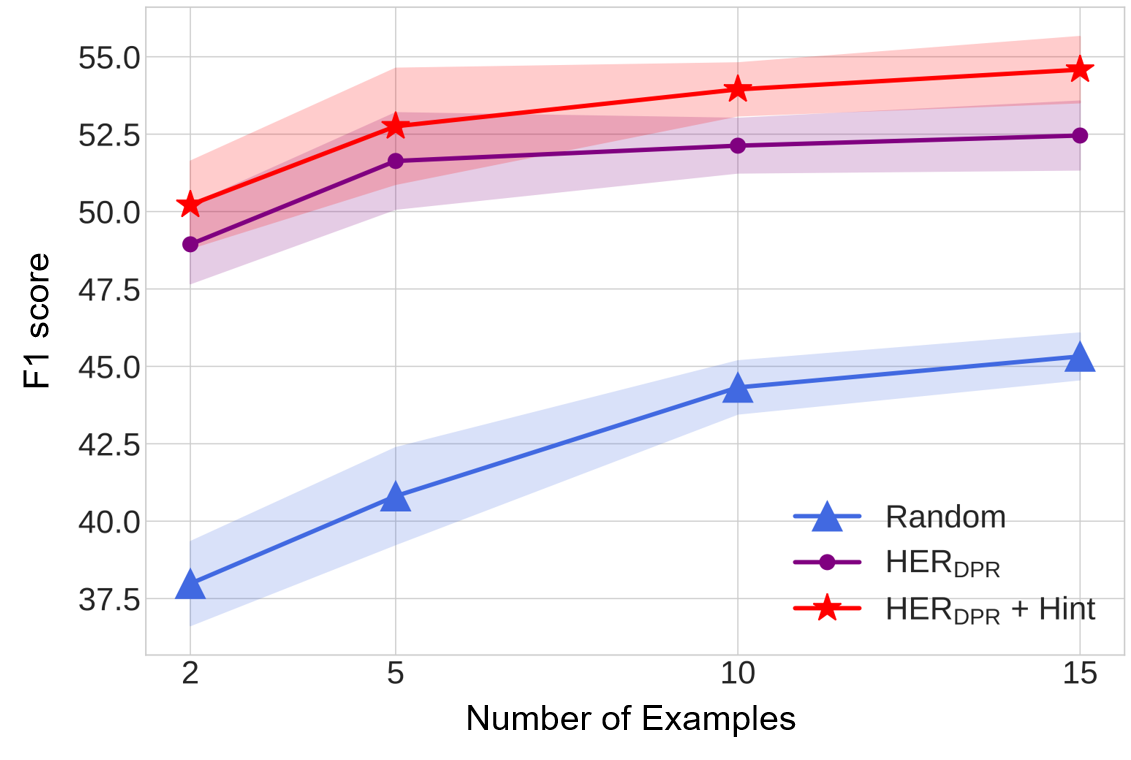}
   \caption{Effect of example numbers for HER$_\mathrm{DPR}$ under standard ICL setting and HICL setting. We report results of gpt-3.5-turbo on NQ dataset.}
   \label{fig:number}
\end{figure}

We investigate the impact of example numbers under the standard ICL setting and HICL setting. We conduct experiments on 2, 5, 10, 15 shots settings and show results of gpt-3.5-turbo in Figure~\ref{fig:number}. 
For the Random selection method, we gain a more significant performance improvement with the increasing number of examples.
The performances of HER$_\mathrm{DPR}$ under the standard ICL and HICL settings both benefit from utilizing more examples. 
Furthermore, HICL brings more performance gains when the number of examples increases, showing that HICL can better attract LLM's attention to the relevant knowledge.

\subsection{Order sensitivity}
Moreover, we explore how example orders may affect results under the standard ICL setting and HICL setting. 
We conduct order sensitivity analysis based on the gpt-3.5-turbo, and select examples with HER$_\mathrm{DPR}$ under 5 shot setting.
For each time, we sample 300 queries from the test set, and the mean scores over 5 random seeds are reported in Table~\ref{tab:order}. The results demonstrate that concatenating the most similar example closest to the query results in optimal performance for both standard ICL and HICL settings, while the reverse shows the opposite results. 
According to the standard deviation, HICL performs more stably robustly with a lower order sensitivity.

\begin{table}
\setlength{\tabcolsep}{11pt}
\centering
\resizebox{0.87\columnwidth}{!}
{
\begin{tabular}{lcccc}
\toprule
&\multicolumn{2}{c}{\textbf{Standard}}& \multicolumn{2}{c}{\textbf{HICL}}\\
\hline
\textbf{Order}& \multicolumn{1}{c}{\textbf{EM}} & \multicolumn{1}{c}{\textbf{F1}} & \multicolumn{1}{c}{\textbf{EM}} & \multicolumn{1}{c}{\textbf{F1}}\\
\hline 
Random$_1$ & 39.80 & 51.34 & 41.60 & 52.44 \\
Random$_2$ & 39.33 & 50.84 & 41.07 & 51.98 \\
\hline 
Reverse & 38.27 & 50.62 & 40.20 & 51.84 \\
Default & 39.87 & 51.63 & 41.53 & 52.75 \\
\hline 
\hline 
STD & 0.64 & 0.40 & 0.56 & 0.36 \\
\bottomrule
\end{tabular}
}
\caption{\label{tab:order}
The order sensitivity analysis on the NQ dataset with HER$_\mathrm{DPR}$. Default indicates concatenating the most similar example closest to the query, while Reverse is the opposite. STD presents the standard deviation of each score.
}
\vspace{-1em}
\end{table}

\section{Conclusion}

In this work, we focus on the ICL ability of LLMs and conduct experiments on open-domain QA as the representative of knowledge-intensive tasks.
We propose a new paradigm called HICL to extract hints from demonstrations, presenting query-related knowledge more explicitly.
To retrieve hint-related examples more effectively, we label examples with hints and further train HERs on different retrievers. 
Our approach achieves consistent performance improvements on open-domain QA and is also applicable for Black-Box LLMs, where the probability distribution is not accessible. 
The improvement brought by HICL benefits from suitable examples in the training set to form demonstrations~\cite{QA_overlap}. 
For broader applications in the real world, several Retrieval-Augmented Generation works \cite{shi2023replug} further improve the quality of retrieved samples from an external corpus. 
We look forward to HICL bringing improvements to future RAG-related work.

\section{Acknowledgments}
This research was partly supported by the National Natural Science Foundation of China (Grant No.U1903213) and the Shenzhen Science and Technology Program (JSGG20220831
093004008).

\vfill\pagebreak

\bibliographystyle{IEEEbib}
\bibliography{strings,refs}

\end{document}